\documentclass{article}
\usepackage{spconf,amsmath,graphicx}
\usepackage{booktabs}
\usepackage{makecell}
\usepackage[T1]{fontenc}
\usepackage{diagbox}
\usepackage{physics}
\usepackage[dvipsnames]{xcolor}
\usepackage{graphicx}
\usepackage{booktabs}
\usepackage{amsmath}
\usepackage{amssymb}
\usepackage{times}
\usepackage{epsfig}
\usepackage{float}
\usepackage{amsmath}
\usepackage{amssymb}
\usepackage{caption}
\usepackage{subcaption}
\usepackage{adjustbox}
\usepackage{makecell}
\usepackage{bm,multirow}
\usepackage{arydshln}
\usepackage[accsupp]{axessibility}
\usepackage[pagebackref,breaklinks,colorlinks]{hyperref}
\usepackage[capitalize]{cleveref}
\usepackage{listings}
\usepackage{xcolor}

\title{Soulstyler: Using Large Language Model to Guide Image Style Transfer for Target Object}
\name{Junhao Chen$^{1,2}$
\qquad Peng Rong$^{1,2}$ 
\qquad Jingbo Sun$^{1,2}$ 
\qquad Chao Li$^{1,2\ast}$ \thanks{*Corresponding author}
\qquad Xiang Li$^{1,2}$ 
\qquad Hongwu Lv$^{1,2}$ }
\address{$^{1}$ College of Computer Science and Technology, Harbin Engineering University, China \\
$^{2}$ Modeling and Emulation in E-Government National Engineering Laboratory, China \\
}

\begin{document}

\twocolumn[{%
\renewcommand\twocolumn[1][]{#1}%
\maketitle
\vspace*{-1cm}
\begin{center}
\centering
\includegraphics[width=0.9\linewidth]{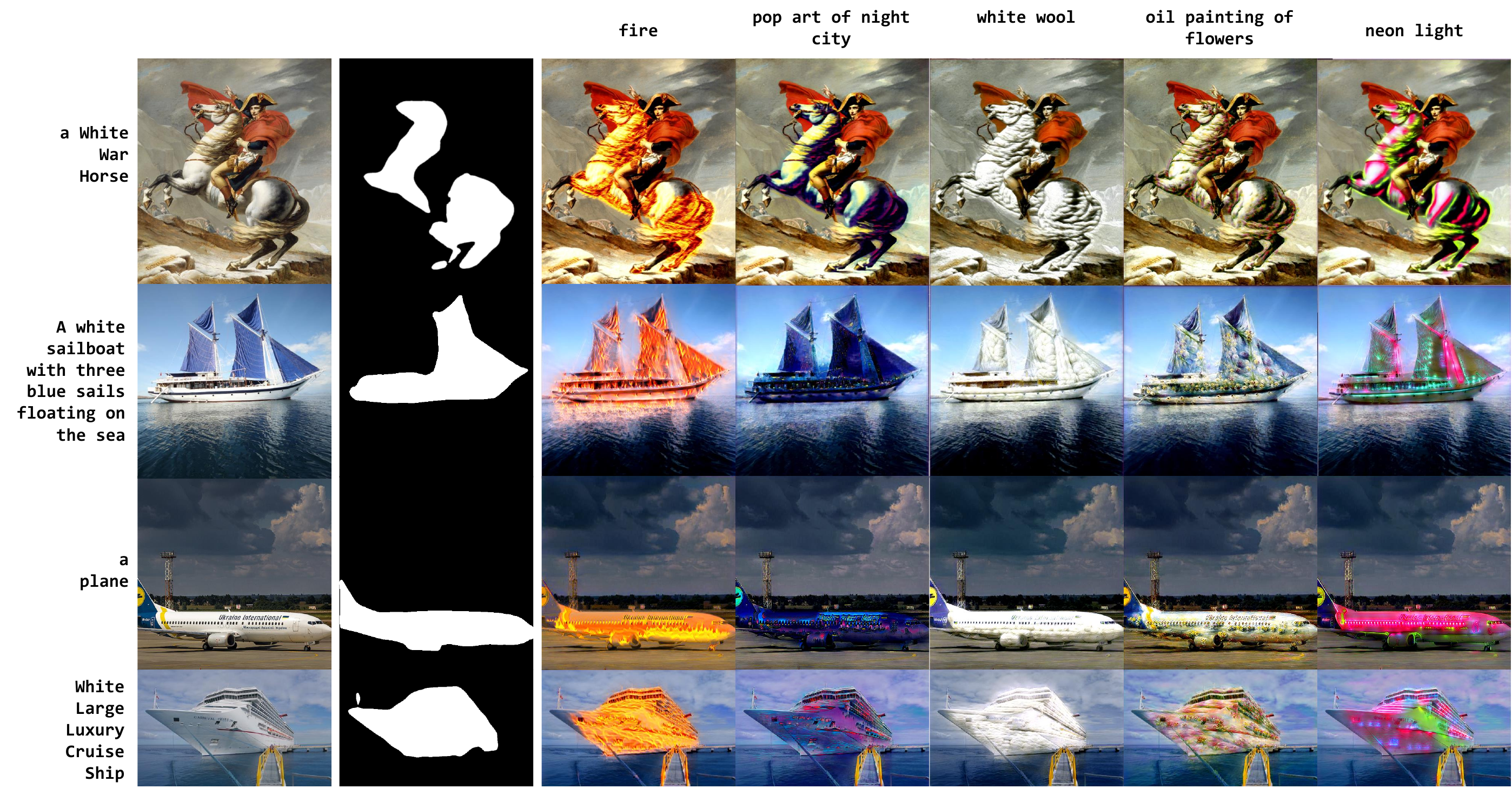}
\captionof{figure}{Our style transfer results on various text conditions. Translated images have spatial structure of the content images with realistic textures corresponding to the text.
}
\label{fig:first}
\end{center}
}]

\begin{abstract}
Image style transfer occupies an important place in both computer graphics and computer vision. However, most current methods require reference to stylized images and cannot individually stylize specific objects.
To overcome this limitation, we propose the Soulstyler framework, which allows users to guide the stylization of specific objects in an image through simple textual descriptions.
We introduce a large language model to parse the text and identify stylization goals and specific styles. Combined with a CLIP-based semantic visual embedding encoder, the model understands and matches text and image content.
We also introduce a novel localized text-image block matching loss that ensures that style transfer is performed only on specified target objects, while non-target regions remain in their original style.
Experimental results demonstrate that our model is able to accurately perform style transfer on target objects according to textual descriptions without affecting the style of background regions.
Our code will be available at https://github.com/yisuanwang/Soulstyler.
\end{abstract}

\begin{keywords}
Image Style Transfer, Target Object, Large Language Model, CLIP, Semantic Visual Embedding
\end{keywords}
\section{Introduction}
\label{sec:intro}

Image style transfer, a key area in computer graphics and computer vision, involves applying artistic styles to images \cite{jing2019neural}. Its applications range from digital media to personalized content creation. However, mastering this technique is challenging due to the complexities in interpreting and applying artistic styles.

Traditional style transfer methods \cite{richardson2021encoding, gatys2016image}, using reference images to transfer styles, often struggle with styling individual objects within an image, especially in applications requiring precise, object-specific stylization. This highlights the need for more accurate and versatile solutions.

Addressing this, we introduce "Soulstyler," a framework that combines the capabilities of large language models (LLMs) like GPT-4 \cite{gpt4} and LLAMA-2 \cite{touvron2023llama} with a CLIP-based semantic visual embedding encoder. Soulstyler facilitates text-guided style transfer, enabling nuanced stylization of specific objects in images.

The framework comprises a text interpretation module powered by an LLM and a visual processing module with a CLIP-based encoder. The language model identifies style attributes and target objects from user-provided text, while the visual encoder precisely applies styles to these objects. A novel localized text-image block matching loss function ensures that only targeted objects are stylized, preserving the original style elsewhere.

Our experiments demonstrate Soulstyler's effectiveness in accurately executing style transfers on specific objects based on textual descriptions. This showcases its potential in diverse fields such as digital art and advertising, underlining its practicality and adaptability.

The study's contributions include the innovative integration of LLMs with visual encoders for targeted style transfer and a unique loss function for preserving the original style in non-targeted areas. These advancements represent significant progress in image style transfer, enhancing its scope and application.

\section{Related Works}
\subsection{Style Transfer}
Neural style transfer has transitioned from VGG-19 network-based pixel optimization \cite{simonyan2014very, gatys2016image} to advanced techniques involving perceptual loss functions, and feature transforms. This progression has improved efficiency, content accuracy, and overall style quality. Further innovations include attention mechanisms, wavelet transform-based methods like WCT \cite{li2017universal, li2018closed}, and graph convolutional networks, enhancing photorealistic transfers and style-content integration.

\subsection{Text-guided Synthesis}
Text-guided image synthesis has evolved significantly, with models such as AttnGAN \cite{xu2018attngan}, ManiGAN \cite{li2020manigan}, CLIP \cite{radford2021learning}, StyleCLIP \cite{patashnik2021styleclip}, CLIPstyler \cite{kwon2022clipstyler}, and StyleGAN-NADA \cite{gal2021stylegan} incorporating advanced attention mechanisms and robust text-image embeddings. Despite their advancements, these models are typically constrained to their trained domains, while our approach offers more flexible, domain-agnostic texture transfers driven by text.

\subsection{Image Semantic Segmentation}
Image semantic segmentation has benefited greatly from deep learning advancements. Starting with foundational models like FCNs and U-Net \cite{ronneberger2015u}, the field has progressed to sophisticated systems like DeepLab \cite{chen2017deeplab}, which employ dilated convolutions and atrous spatial pyramid pooling. Recent approaches, notably CRIS \cite{wang2022cris}, have effectively used text-image embeddings, particularly from CLIP, for enhanced segmentation accuracy.

\subsection{Large Language Models}
LLMs are pivotal in natural language processing, achieving remarkable language understanding and generation capabilities through extensive pre-training \cite{chatgpt, gpt4, touvron2023llama}. Their versatility extends to multimodal applications, including image and video processing, where they enable effective cross-modality interactions.

\section{Method}
\subsection{Overall Architecture}
\begin{figure}[htbp]
\centering
\small
\includegraphics[width=0.45\textwidth]{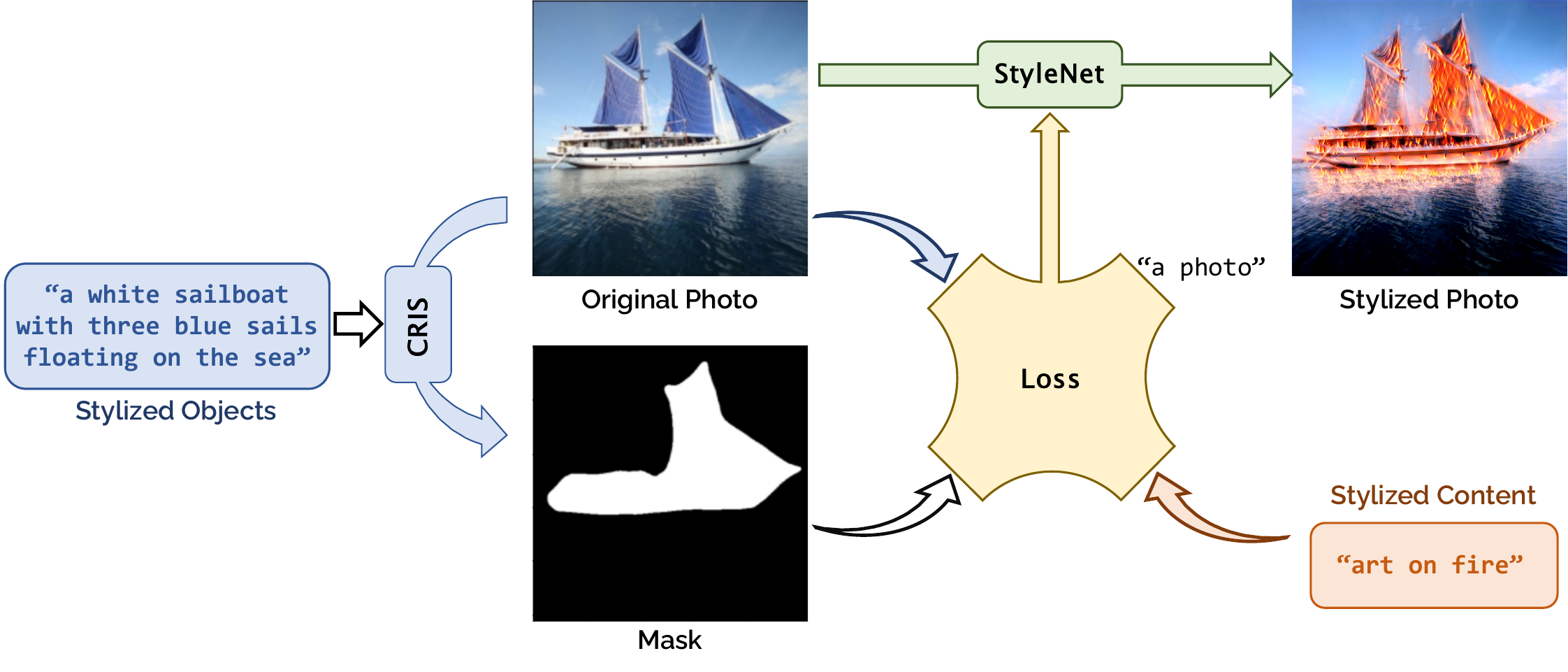}
\caption{The overall architecture of the system.}
\label{fig:soulstructure}
\end{figure}

The overview of the system is shown in Figure \ref{fig:soulstructure}. The overall architecture of our network is designed to transfer the style from a given stylized instruction to a content image, $f(I_o)$, resulting in a stylized output image, $I_s$. This is achieved using a CNN encoder-decoder model, $f(StyleNet)$, which is capable of capturing the visual features of the content image and stylizing the image in the deep feature space to obtain a realistic texture representation. Hence, the stylized image $I_s$ is represented as $f(I_o)$, and the ultimate goal is to optimize the parameters of $f$.

A significant innovation in our work lies in the modification of the loss function. Building upon the foundation of CLIPstyler \cite{kwon2022clipstyler}, we incorporate a mask layer from the CRIS \cite{wang2022cris} model into our loss function, thereby controlling the text feature and image feature losses in the StyleNet, and ultimately achieving style transfer effects in the specified region.
The loss function can be expressed as:
\begin{equation}
L_{\text{total}} = \lambda_d L_{\text{dir}} + \lambda_p L_{\text{patch}} + \lambda_c L_c + \lambda_{\text{tv}} L_{\text{tv}} + t \lambda_m L_{\text{mask}}
\end{equation}
The loss function includes several components. $L_{\text{total}}$ represents the total loss, $L_{\text{dir}}$ is the directional CLIP loss, $L_{\text{patch}}$ is the patchwise CLIP loss, $L_c$ is the content loss, $L_{\text{tv}}$ is the total variation regularization loss, and $L_{\text{mask}}$ is the mask loss from CRIS. The weights for each of these losses are represented by $\lambda_d$, $\lambda_p$, $\lambda_c$, $\lambda_{\text{tv}}$, and $\lambda_m$ respectively. Additionally, $t$ is a threshold that controls whether stylization is performed in the mask region; in this paper, it is set to 0.7.

The mask loss, $L_{\text{mask}}$, is crucial as it ensures that the style transfer is applied specifically to the regions of interest defined by the CRIS-generated mask. This results in a more precise and controlled style transfer, which is particularly beneficial for applications requiring targeted stylization.

\subsection{LLM Prompt Engineering}
Prompt engineering plays a crucial role in our method as it involves transforming the input stylized instruction into separate stylized content and stylized objects. For this task, we experimented with various 10-billion parameter scale open-source language models, including ChatGLM-6B \cite{du2021glm}, ChatGLM2-6B \cite{du2021glm}, BELLE-7B \cite{belle2023exploring}, Baichuan-7B \cite{baichuan}, ChatFlow \cite{li2022csl, zhao2022tencentpretrain}, Phoenix-Inst-Chat-7B \cite{chen2023phoenix}, ChatYuan-large-v2 \cite{clueai2023chatyuan}, Moss-Moon-003-SFT \cite{sun2023moss}, RWKV \cite{peng_bo_2021_5196578} and Llama 2-7B \cite{touvron2023llama}.

\begin{figure}[htbp]
\centering
\small
\includegraphics[width=0.4\textwidth]{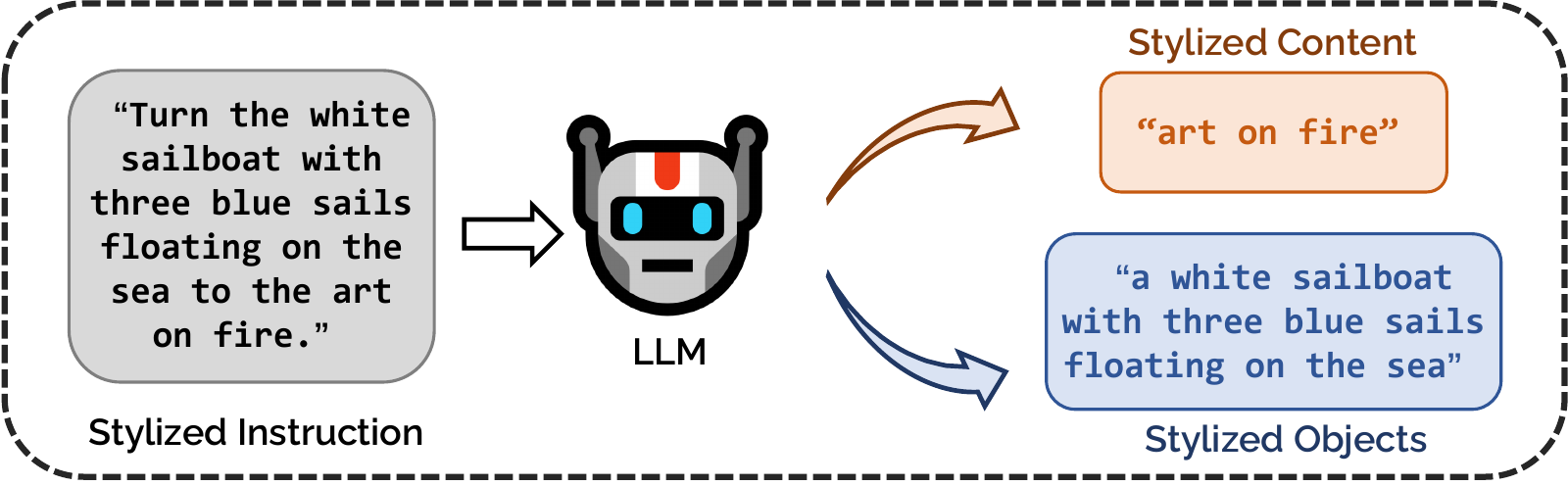}
\caption{Splitting Stylized Instruction into Stylized Content and Stylized Objects using the LLM.}
\label{fig:soulllm}
\end{figure}

Each of these models was tasked with splitting the stylized instruction into its constituent components of stylized content and stylized objects. This step is crucial as it sets the stage for the subsequent image style transfer process. Moreover, the efficiency and accuracy of this step can have a significant impact on the final results. As shown in Figure \ref{fig:soulllm}.

To evaluate the performance of each model in segmenting the stylized instruction, we employed ChatGPT \cite{chatgpt} as a benchmark for assessing the segmentation results produced by each LLM. This involved scoring the output of each model based on its ability to accurately and meaningfully segment the stylized instruction. This assessment was necessary to determine the most suitable model for our application and to ensure the highest quality of the final stylized image.

\subsection{Basic Framework of CLIPstyler}
CLIPstyler \cite{kwon2022clipstyler} leverages the CLIP model to transfer the semantic style of a target text to a content image without requiring a specific style image. It uses the StyleNet model to capture and style visual features in deep feature space, and optimizes it using a combined loss function as proposed by CLIPstyler. This loss function includes the total loss, directional CLIP loss, patch CLIP loss, and total change regularization loss, with respective weights, enabling the optimization of parameters for generating stylized images without a specific reference image.

\subsection{CRIS for Semantic Segmentation of Images}
The CLIP-Driven Referring Image Segmentation (CRIS) \cite{wang2022cris} framework is utilized in our work for performing semantic segmentation of images, which is crucial for generating binary mask images.

\begin{table}[h]
\centering
\tabcolsep=0.08cm
\caption{Segmentation score of different LLMs. We performed a thorough manual evaluation of the 100 manually set stylization commands and the corresponding standard answers (stylized content and stylized objects). LLM outputs are marked as correct when the stylized content and stylized objects are in perfect agreement with the standard answers, and the right-most columns are the scores we got from manually evaluating the segmentation effects of the LLMs.}
\label{table:llm_segmentation_score}
\begin{tabular}{lccc}
\hline
Model & \begin{tabular}[c]{@{}c@{}}GPT3.5\\Score\end{tabular} & \begin{tabular}[c]{@{}c@{}}GPT-4\\Score\end{tabular}   &\begin{tabular}[c]{@{}c@{}}Manual\\Evaluation\end{tabular}\\
\hline
ChatGLM-6B \cite{du2021glm} & 6.84 & 6.94  &51\%\\ 
ChatGLM2-6B \cite{du2021glm} & 9.38 & 9.48  &77\%\\ 
BELLE-7B \cite{belle2023exploring} & 4.32 & 4.30  &21\%\\ 
Baichuan-7B \cite{baichuan} & 2.88 & 3.09  &27\%\\ 
ChatFlow \cite{li2022csl, zhao2022tencentpretrain} & 4.28 & 4.52  &43\%\\ 
Phoenix-Inst-Chat-7B \cite{chen2023phoenix} & 6.30 & 6.29  &54\%\\ 
ChatYuan-large-v2 \cite{clueai2023chatyuan} & 2.51 & 2.43  &18\%\\ 
Moss-Moon-003-SFT \cite{sun2023moss} & 4.29 & 5.12  &69\%\\ 
RWKV \cite{peng_bo_2021_5196578} & 5.90 & 5.42  &61\%\\
Llama 2-7B \cite{touvron2023llama} & 9.62 & 9.23  &84\%\\
\hline
\end{tabular}
\end{table}

\section{Experiments}
\subsection{Prompt Engineering}

Prompt engineering plays a pivotal role in our style transfer process. It involves crafting precise prompts that guide a Large Language Model (LLM) to segment a stylized instruction into two components: stylized content and stylized objects. This segmentation is critical, as it defines the specific areas and styles to be applied in the content image.

To assess the effectiveness of various LLMs in prompt engineering, we conducted experiments with multiple 10B parameter-level models. These included ChatGLM-6B \cite{du2021glm}, ChatGLM2-6B \cite{du2021glm}, BELLE-7B \cite{belle2023exploring}, Baichuan-7B \cite{baichuan}, ChatFlow \cite{li2022csl, zhao2022tencentpretrain}, Phoenix-Inst-Chat-7B \cite{chen2023phoenix}, ChatYuan-large-v2 \cite{clueai2023chatyuan}, Moss-Moon-003-SFT \cite{sun2023moss}, RWKV \cite{peng_bo_2021_5196578}, and Llama 2-7B \cite{touvron2023llama}. We used ChatGPT \cite{chatgpt} as a benchmark to compare each model's segmentation capability, evaluating their performance based on the clarity and accuracy of the generated instructions.

We use the default prompt:
\textit{Split ["Turn the white sailboat with three blue sails floating on the sea to the art on fire."] into [Stylized Content] and [Stylized Objects]. Returns a json with two keys: StylizedContent and StylizedObjects.}

A sample response:
\textit{\{
 "Stylized Content": "art on fire",
 "Stylized Objects": "the white sailboat with three blue sails floating on the sea"
\}}

\subsection{Selection of Large Language Model}

Selecting an appropriate LLM is essential for accurately interpreting stylized instructions and guiding the style transfer process. We evaluated several models based on their ability to segment stylized instructions and the quality of the resulting stylized content and objects. Llama 2-7B \cite{touvron2023llama} and ChatGLM2-6B \cite{du2021glm} performed excellently, as shown in Table \ref{table:llm_segmentation_score}, but we ultimately chose Llama 2-7B for our application.

\subsection{Selection of Stylization Threshold}

\begin{figure}[htbp]
\centering
\small
 \includegraphics[width=0.45\textwidth]{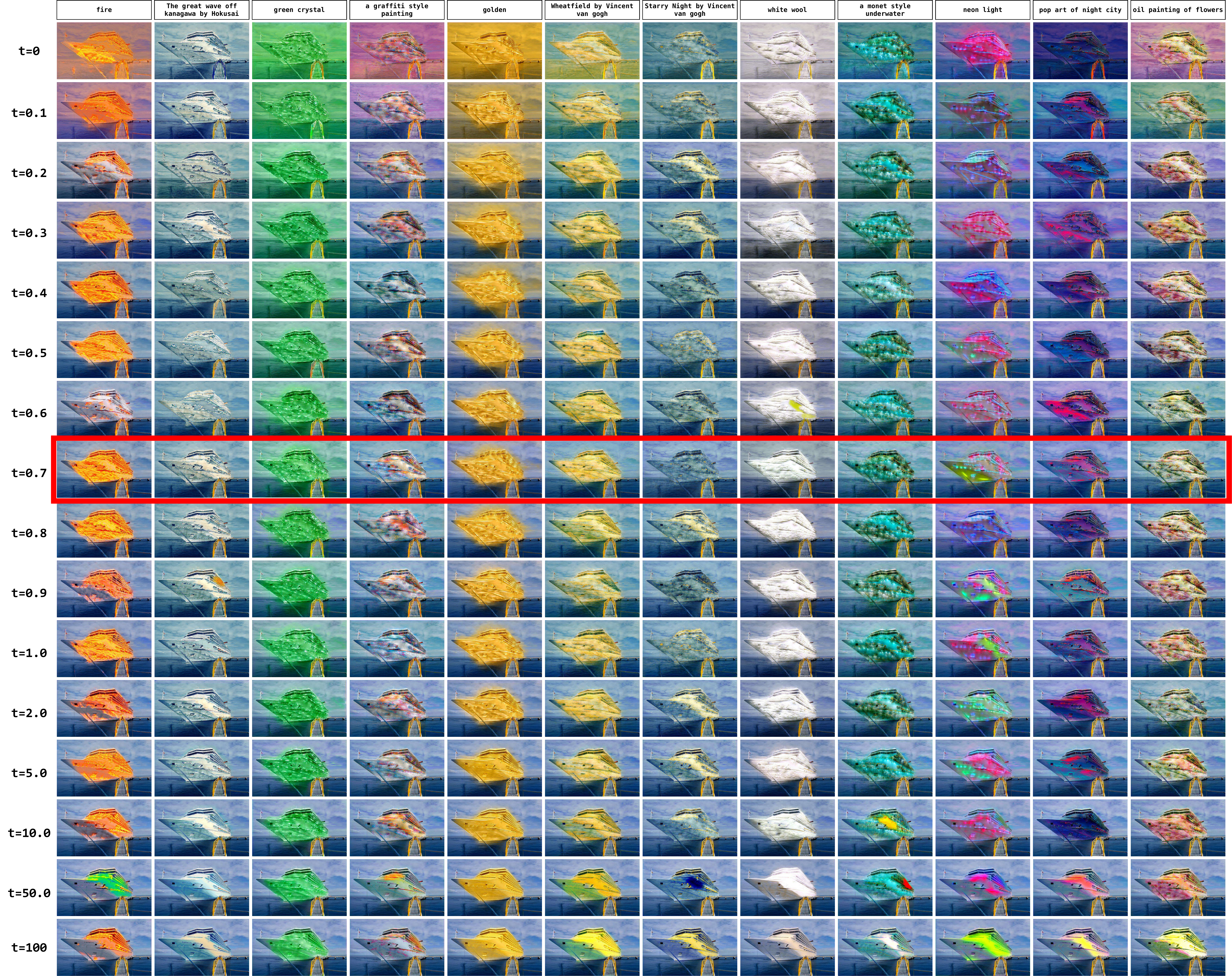}
\caption{Experiments with varying stylization thresholds. The threshold $t = 0.7$ demonstrates an optimal balance between stylization and original image features.}
\label{fig:expt}
\end{figure}

Selecting the appropriate stylization threshold is crucial for achieving a balance between distinct style transfer and content preservation. We determined that a threshold of $t=0.7$ optimally balances these aspects. It effectively harmonizes the stylization of the targeted object's unsegmented areas with the original features of the non-target regions, maintaining a coherent blend of style, texture, and color. Comparative experiments with varying threshold levels are illustrated in Figure \ref{fig:expt}, which showcases the impact of different threshold settings on the stylization process.

\begin{figure}[htbp]
\centering
\small
\includegraphics[width=0.45\textwidth]{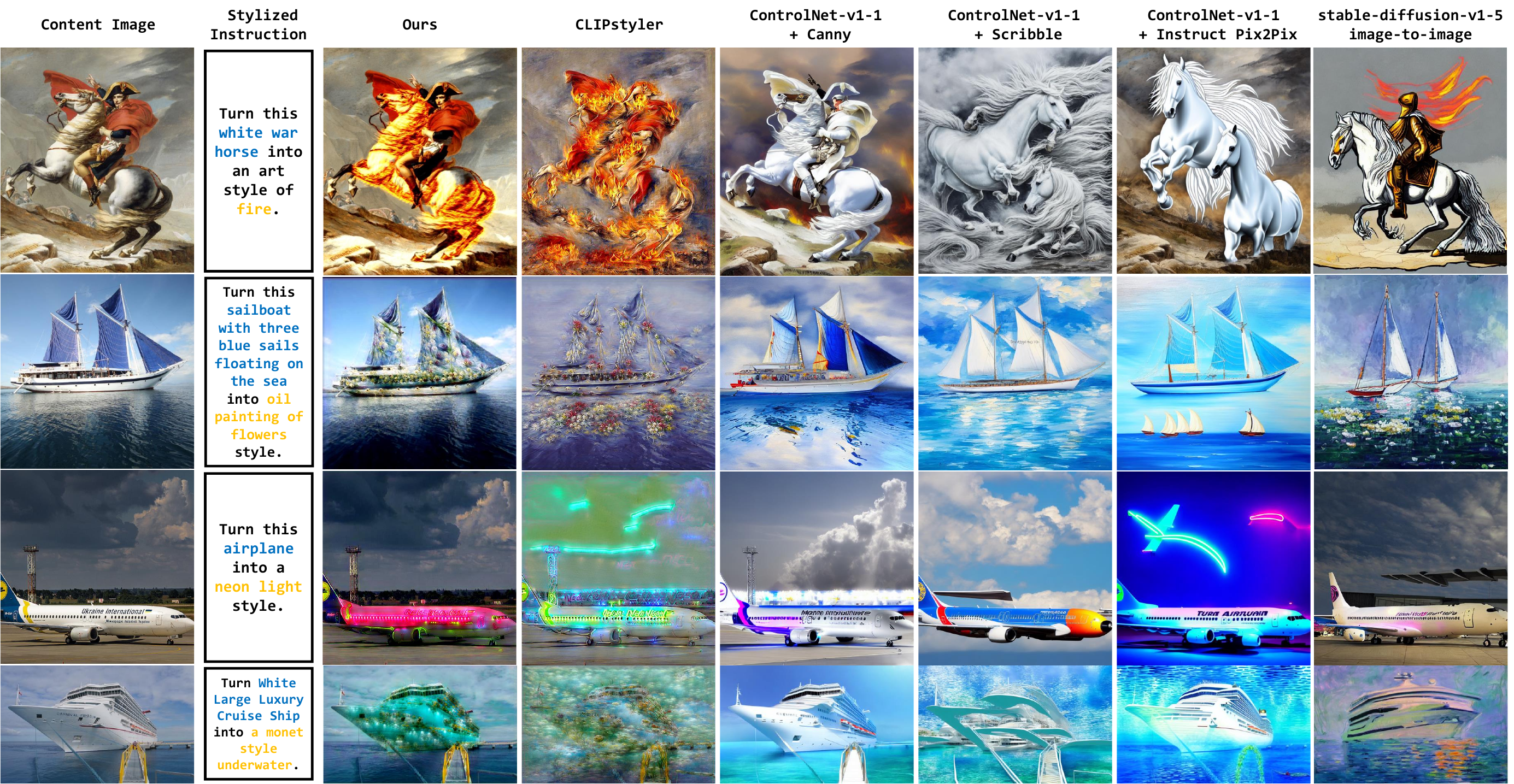}
\caption{Comparison with leading text-guided image style transfer models, including ControlNet \cite{zhang2023adding}, CLIPstyler \cite{kwon2022clipstyler}, stable-diffusion-v1-5 \cite{Rombach_2022_CVPR}. CLIPstyler and stable-diffusion-v1-5 are shown alongside other baselines, with a focus on how they interpret style instructions. For a fair comparison, images from models that output square images are adjusted to match the input Content Image's aspect ratio.
}
\label{fig:comp}
\end{figure}

\section{Results}
We tested our style transfer method on a large number of samples to evaluate its performance, as shown in Figure \ref{fig:first}. The figure shows the results of applying our method to test images with different contents and styles, indicating that our method successfully transfers styles from the target text to the content images while preserving the original contents. This demonstrates the versatility and robustness of our method in dealing with various styles and contents. Ultimately, the results show that our proposed method effectively achieves realistic and visually appealing style transfer while preserving the original image content. The chosen model, threshold, and CRIS for semantic segmentation play a key role in the success of the method, as evidenced by the high-quality test results. More experimental results can be viewed by going to the repository of this project.
Our Soulstyler has the best visuals, as shown in Figure \ref{fig:comp}, remarkable success in image consistency, and the ability to fully satisfy input stylization commands.

\section{Conclusion}

This study introduces a revolutionary approach to controlled style transfer, overcoming existing challenges and adding new features to improve the stylization process's quality and controllability. Utilizing LLMs for prompt engineering and integrating the CLIP-Driven Referring Image Segmentation (CRIS) framework, we have devised a method that enables controlled stylization regions, text-based style descriptions, and preservation of original content. Extensive experiments confirm our approach's effectiveness in producing visually appealing results while preserving the original image content. The integration of LLMs and CRIS, along with the optimal stylization threshold, makes our method one of the most advanced controlled style transfer solutions available.
Our approach opens up new possibilities in art, design, and other creative fields, offering artists and designers more control over the stylization process and fostering creativity in innovative ways. We believe our method is a valuable addition to the controlled style transfer field and may inspire further research and innovation.



\vfill\pagebreak

\ninept
\bibliographystyle{IEEEbib}
\bibliography{strings,refs}

\end{document}